# Research on the Mobile Robots Intelligent Path Planning Based on Ant colony algorithm Application in Manufacturing Logistics

GUO Yue [1], SHEN Xuelian [1],ZHU Zhanfeng[1]
Management Engineering Institute, Ningbo University of Technology, No.201 Fenghua Road, Ningbo. 315211 Zhejiang, China
guoyue@nbut.edu.cn; shenxuelian@nbut.edu.cn; zhuzhanfeng@nbut.edu.cn

***Abstract***

*With the development of robotics and artificial intelligence field unceasingly thorough, path planning as an important field of robot calculation has been widespread concern. This paper analyzes the current development of robot and path planning algorithm and focuses on the advantages and disadvantages of the traditional intelligent path planning as well as the path planning. The problem of mobile robot path planning is studied by using ant colony algorithm, and it also provides some solving methods.*

## 1. Introduction

The research of mobile robot started from the late 1960s.The Stanford Institute successfully developed the autonomous mobile robot—Shakey robot in 1966, The robot has independent reasoning, planning, control and other functions in complex conditions with the application of artificial intelligence. At the end of the 1970's, the application of computer and sensor technology researches on mobile robot reach to a new high tide as a result of the development. The mid 1980's, a large number of world famous company started to develop mobile robot platform. The mobile robot is mainly used as the mobile robot experiment platform in university laboratories and research institutions, and promoting the multi-directional learning of the mobile robot. Since the 1990s, the symbol of environment information sensor and information processing technology development of high level, high adaptability of mobile robot control technology, programming technology under the real environment has emerged, and the higher level research of mobile robotics able to be conducted. In recent years, mobile robots are widely used in space exploration, ocean development, atomic energy, factory automation, construction, mining, agriculture, military, and service, etc. Research on mobile robot has become a hot research issue and the concern of the international robot.

Intelligent mobile robot is a set of integrated system of multiple functions which consists of environment perception, dynamic decision-making and planning, behavior controlling and executing. In recent years, mobile robot has wide application prospect in space exploration, ocean development, atomic energy, factory automation, construction, mining, agriculture, military, and service, etc. China started the research on the intelligent robots later than some developed countries, and there still existed a big gap within China and developed countries. In recent years, the research theory and method for robot have reached the international advanced level by the China Robotics Lab, and achieved a number of important scientific research achievements in robotics frontier exploration and demonstration application etc. Because the state and society paid much attention on the robot field, which including: all-weather 120Kg suspended wing UAV system, polar research snow mobile robot, Ling Lizard-anti-terrorism and anti-riot robot, robot nano operating system, etc.

Mobile robot's path programming technology is one of the core technology in the field of robot research, which study of the algorithms is advantageous to the improvement of robot planning to meet the needs of practical applications. The path programming is that, in the obstacle environment, according to a certain evaluation standard, finding collision free path from the initial state to the target state. The main issues include finding the optimal or approximate optimal one from the initial state to the target state collision through the free path and an algorithm to built reasonable model by using of the mobile robot environmental information. In the model which being able to cope with uncertain factors and path tracking errors in the environment, making the influence of external objects to the robot reduced to a minimum: how to use all the information known to guide the robot motion, resulting in a better decision ［1］.

## 2. LITERATURE REVIEW

The traditional method of path programming is carried out simulation test based on graph. The general approach is based on the global path planning. At present domestic and foreign common methods include grid method, topology, visibility graph, Voronoi graph, method, the artificial potential field method, A* algorithm etc.

Grid method proposed by Howden in 1968 ［2］, decomposing the robot planning space into a number of information network unit working space is divided into unit after the use of heuristic algorithm to search the safe path in the unit. ［3］ The search process always uses work space with four quadtree or octree. Consistency and standards grid makes simple adjacency relation in raster space. After giving each grid traffic factor, path planning problem turns into a problem of searching optimal path with two grid nodes in the grid network. Topological method is mainly divided space with topological feature subspace, and then look for topological path is the starting point to the target point based on the topology of the network, and finally find the path geometry by the topological path.[4] The basic idea of the method is to find paths in high dimensional space transformed into the problem of determining the connectivity of the problem of Low Dimensional Topology space.[5] The visibility graph method is a kind of configuration space method,[6] it mainly regards the robot as a particle processing, expanding the boundary of the corresponding outward obstacles in the work environment, and the boundary is formed with vertices of polygons, determining its vertices, including the robot starting point and the target point. These points connect, but each vertex cannot connect across each other, forming a visibility graph.

The Voronoi plot method is first discovered by the Russian mathematician Voronoi which can be applied to static random environment, which is to say in the process of robot running, the environment is static.[7] All the obstacles are motionless, but environment is uncertain before the robot starts its path planning, the size and the location parameter of the obstacles in the environment is changeable. This method using the path which may be far away from obstacles to show walls arc, that results the path will increase from the initial node to the target node.

Artificial potential field method is a method of local path planning, proposed by Khatib etc.[8] The basic idea of the method is regarding the motion of the robot in the environment as a virtual artificial force field motion.[9] Obstacles generate repulsive force on the robot, and attraction on the target. The joint force of attraction and repulsive force controls robot motion direction, which will determine the position of the robot[10].

In recent years, with the rapid development of in-depth research and the modern computing technology in mobile robot path programming, the traditional path programming is hard to meet its requirements and failure in meeting the need of actual environment changes. Therefore intelligent path algorithms have been studied and used in robot path programming widely. The artificial intelligent path programming algorithm improves the accuracy of robot obstacle avoidance path programming greatly and accelerates the programming speed, all these are to be met the needs of practical application. Intelligent path programming algorithm includes genetic algorithm[11][12], particle swarm algorithm[13], fuzzy logic[14][15], neural network[16][17], artificial immune algorithm[18] and hybrid algorithm[19][20][21]. The above algorithms have been made certain achievements for the robot protecting the obstacle in known or unknown circumstances.

Italian scholar Dorigo and Colorni proposed a heuristic optimization algorithm in 1991, which is biologically inspired.[22] It simulated and reference the behavior of ants in the real world to solve combinatorial optimization problems under distributed environment.[23] It also solves the problems of large cost when robot in complex environment contains a large number of irregular obstacles in the path programming[24].

Ant colony algorithm is produced to simulate the process of ants foraging. Ants release specials in the search of path when confronted with a no through road, they will randomly select one while releasing hormone information of path length. When the ants again encountered this intersection, optimal path on the pheromone concentration increase, while the other pheromone concentration is cutting with the passage of time.[25] At the same time, the ants can adapt to changes in the environment when obstacles emerge, they will find an optimal path to go. Ant colony algorithm has the features of group cooperation, positive feedback and distributed computing. Group cooperation is a cooperation for better optimization task. Although each artificial ant can build a solution, but the solution with high quality is always produced by ant colony cooperation. The feedback mode of the algorithm is used in the optimum solution which leaves more pheromone on a path, and more pheromone in turn attracts more ants. The positive feedback process guides the system towards the optimal solution of the evolution direction. Distribute Computing of Ant colony algorithm can calculate each artificial ant at the multiple points in the problem space, at the same time, it began to separate the structural problems of solutions. The result will not be affected only because of one artificial ant cannot successfully obtained the solution Distributed Computing makes the algorithm easy to be implemented.[26] These characteristics make the ant colony algorithm suitable for solving complex combinatorial optimization problems.

The path programming is a kind of combinatorial optimization problem, so the ant colony algorithm is suitable to solve the path planning problems.

## 3 QUESTIONS

### 3.1 Problems in design

In a 800 × 800 planar scene graph, there is a robot at the origin of O(0, 0), which can only activities in the planar scene range. The 12 different shapes of the regions are the obstacles that the robot cannot collide The description of the mathematical are as shown in the following table:

Table 1. Planar scene graph description

| **Region No.** | **Obstacles' name** | **vertex coordinates at left corner** | **Other description of the characteristics** |
|---|---|---|---|
| 1 | square | (300, 400) | length of 200 |
| 2 | circular | | center coordinates (550, 450) a radius of 70 |
| 3 | parallelogram | (360, 240) | Base length140, the coordinates of the vertices on the left top(400, 330) |
| 4 | triangles | (280, 100) | top vertex coordinates (345, 210)， the lower right vertex coordinates (410, 100) |
| 5 | square | (80, 60) | length 150 |
| 6 | triangles | (60, 300) | top vertex coordinates (150, 435)， the lower right vertex coordinates(235, 300) |
| 7 | rectangular | (0, 470) | length 220， width 60 |
| 8 | parallelogram | (150, 600) | Base length 90， the coordinates of the vertices on the left top(180, 680) |
| 9 | rectangular | (370, 680) | length60， width120 |
| 10 | square | (540, 600) | length130 |
| 11 | square | (640, 520) | length80 |
| 12 | rectangular | (500, 140) | length300， width60 |

### 3.2 Maintaining the Integrity of the Specifications

Specify a point outside the obstacles as the target for the robot to reach (target point and the distance to the obstacle are at least more than 10 units). Set the rule thar walking path of the robot should by lines and arcs. The arc parts are robot's turning path. A robot cannot turn by line. The turning path consists with straight path tangent to a circle, and can also be composed of two or more circular arc path, but the minimum radius of each circular arc path is 10 units in order not to collide with obstacles. It also requires the distance between the robot walking route and obstacle is no more than 10 units, or a collision will occur and the robot can not complete the straight walking. 4 point O in the scene graph(0, 0), A(300, 300), B(100, 700), C(700, 640).

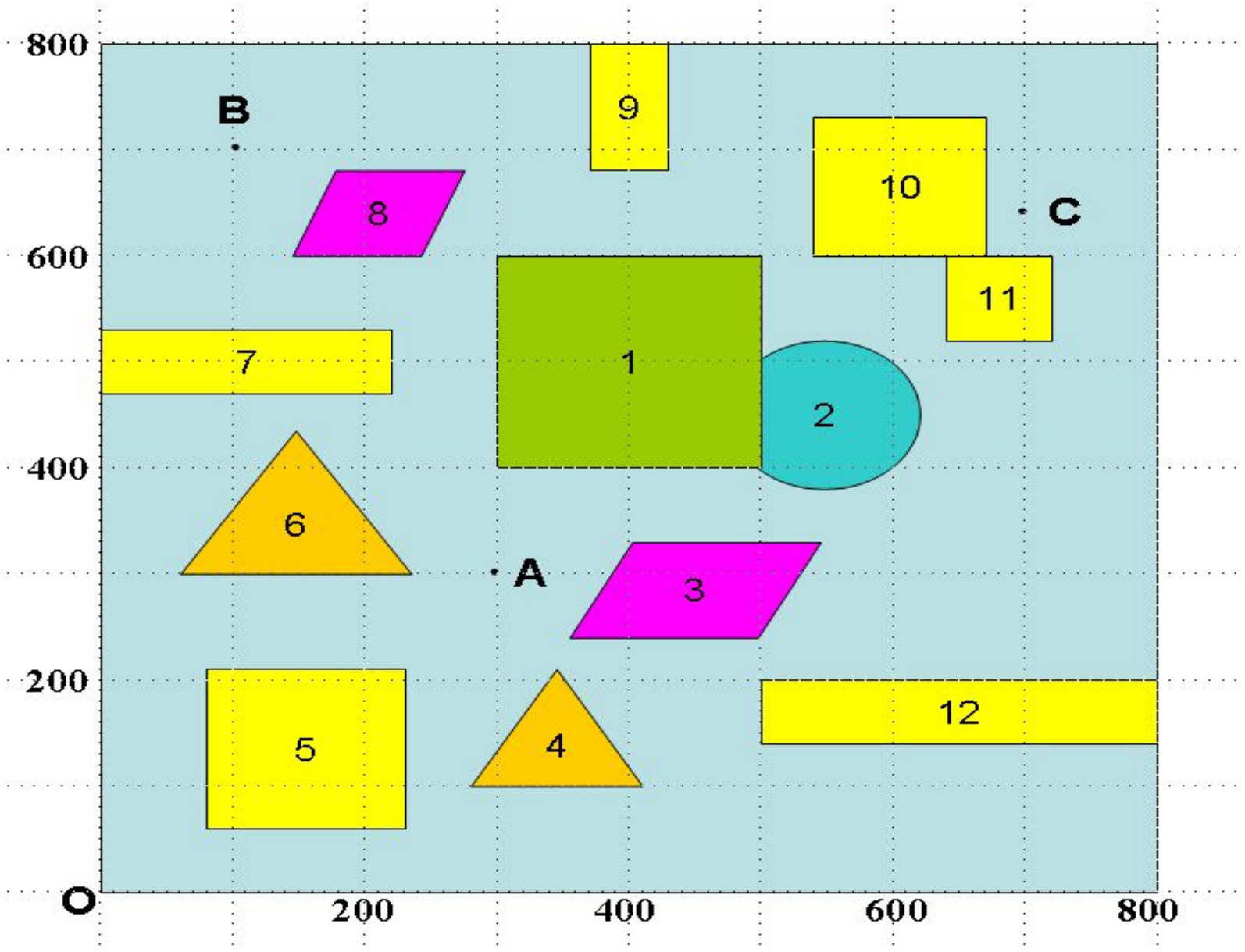

Fig1. 800 × 800 planar scene graph

The maximum speed of the robot walking straight into 5 unit / sec. and maximum turning speed is

$$v = v(\rho) = \frac{v_0}{1 + e^{10 - 0.1\rho^2}}$$

where P is radius of turn . If the speed is higher than that, the robot will rollover, and is unable to walk.

**Question: Start from O (0, 0) ,which is the shortest path of O → A, O → B ?**

# 4 METHODOLOGY

In order to find the shortest path to the target from (0,0) with certain rules walking around obstacles ,we can draw the envelope of robot walking hazardous area, just around the corner with a radius of 10 units a quarter arc, by the method of the rope then to find the shortest possible path (for example, seek the shortest path between O and A, can be connected to the section of rope between O and A, to the arc of the corner support taut, then the length of this wire is O to a shortest possible path (A),and then list shortest path possible paths to each target point with Brute-force method.

Designated O (0,0) after the middle of a number of points around obstacles to reach the target point according to certain rules in the back of O, which allows us to consider not just obstacles inflection point, should be considered after the target point in the path at the turn of the problem.Simple line circle structure can not solve this problem,so we have adopted the form of a minimum turning radius at the inflection point and the target point on the way.We can also be appropriate to transform the inflection point of the turning radius, so that the robot can along straight line through the target point of the way, and then create optimization model to optimize these two programs, and ultimately obtained the shortest path.

The model assumes and symbol description are analyzed by the following assumptions:

(1)Assuming the width of the robot itself is negligible. Thus, the movement of the robot can be regarded as a point moves.

(2)Assuming the robot walk straight and turn at maximum speed.

(3)Assuming that the obstacle is always subject to 12 different shapes of the area and the nature of the location, size, etc. has been the same.

**Table2 Symbol & Description**

| Symbol | Description |
|---|---|
| $v_0$ | Maximum speed straight line when walking |
| $\rho$ | Turning radius |
| $S_i$ | The i-th sub-length of the shortest path from $O \to A$ |
| $L_i$ | The i-th sub-length of the shortest path from $O \to B$ |
| $L_i^{'}$ | The i-th sub-length of the shortest path from $O \to C$ |
| $t_{\min}$ | The shortest time from $O \to A$ |
| $\widehat{B_iB_j}$ | Arc length from $B_i \to B_j$ |
| $\lvert B_iB_j \rvert$ | Length of the line from $B_i \to B_j$ |

# 5 FINDINGS AND INTERPRETATIONS

## 5.1 The shortest path from $O \to A$ optimization model

Two points within the plane of the shortest path based on the length of the line segment as the endpoint,but the connection of these two segments with obstacles intersect, so try to attempt to bypass the obstacle and its hazardous areas other path.Obstacle is a square, the center of this square is located in the lower part of the connection, so the robot to bypass the obstacle from the top of the obstacle path is the shortest path.

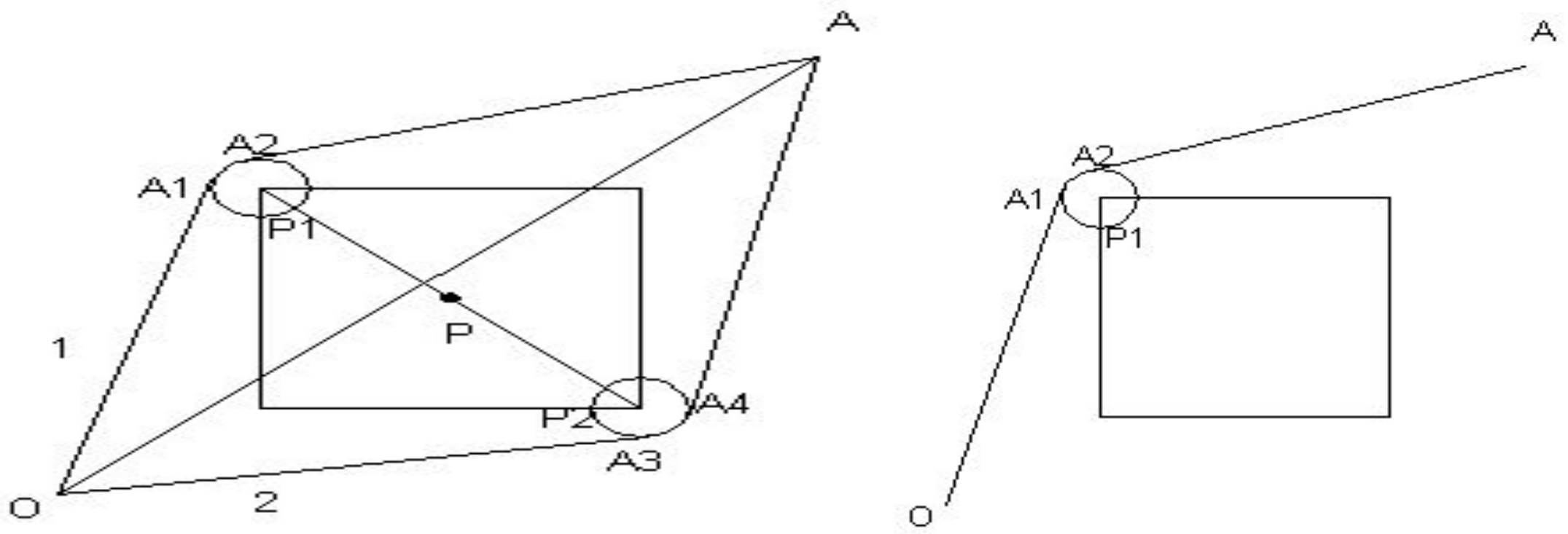

Figure 2. $O \to A$ path

Shown in Figure 2, the shortest path from $O \to A$ is constituted by straight line $OA_1$ and $A_2A$ and a tangent arc $\widehat{A_1A_2}$ wherein the cut point. Which $A_1$ $A_2$ as the cutoff point.Arc $\widehat{A_1A_2}$ thought the center of the circle $P_1(80,210)$ ,the radius is 10. Set cut-point coordinates $A_1(x_1,y_1)$ $A_2(x_2,y_2)$ ,these three sections of the path length can be calculated:

$$A_2A = s_3 = \sqrt{(x_2 - 300)^2 + (y_2 - 300)^2} \quad OA_1 = s_1 \tag{1}$$

$$OA_1 = s_1 = \sqrt{(x_1 - 0)^2 + (y_1 - 0)^2} \tag{2}$$

$$\widehat{A_1A_2} = s_2 = 2arc\sin\frac{\sqrt{(x_1 - x_2)^2 + (y_1 - y_2)^2}}{20} \cdot r \tag{3}$$

Here we have $x_1, y_1$, $x_2, y_2$ for the decision variables, the total length of the shortest path as the objective function: Min $s = \sum_{s_1}^{3} s_i$

Constraints: radius $r = 10$, $OA_1^2 + A_1P^2 = OP^2$, $A_2A_1^2 + PA_1^2 = A_2P$ ,points $A_1, A_2$ on the arc.In summary, the structure optimization model as follows:

$$\text{Min } s = \sum_{s_1}^{3} s_i = \sqrt{x_1^2 + y_1^2} + \sqrt{(x_1 - 300)^2 + (y_2 - 300)^2} + 20t \tag{4}$$

$$s.t.\begin{cases} x_1 \le 80 \\ x_1 \ge 70 \\ y_1 \le 220 \\ y_1 - \sqrt{100 - (x_1 - 80)^2} = 210 \\ x_1^2 + y_1^2 + 100 = 50500 \\ (x_1 - 300)^2 + (y_2 - 300)^2 + 100 = 56500 \\ \sin t = \dfrac{\sqrt{(x_1 - x_2)^2 + (y_1 - y_2)^2}}{20} \end{cases} \tag{5}$$

Solving the above model (see Appendix 1 lingo procedures), the results are as follows:

1) The $OA$ length of the shortest path: $s_{\min} = 471.0372$

2) Two arc tangent point coordinates: $A_1 = (70.50596, 213.1406)$ $A_2 = (76.60640, 219.4066)$

Robot shortest path from point O to point A is reached can be expressed in the table3:

Table3. the shortest path from $O \to A$

| No | Start | End | Types of segments | Length |
|---|---|---|---|---|
| 1 | （0，0） | (70.50596,213.1406) | Straight line | 224.4994 |
| 2 | (70.50596,213.1406) | (70.6064,219.4066) | (80,210) as the center of the arc | 9.1105 |
| 3 | (70.6064,219.4066) | （300，300） | Straight line | 237.4273 |
| Total length | | | | 471.0372 |

## 5.2 The shortest path from $O \rightarrow B$ model

In this section, we will simplify the roadmap as an empowered network diagram, and use the ant colony algorithm to find the approximate route of the shortest path.

The ant colony algorithm is a bionic algorithm derived from the nature of ants routing mode simulation . Ants in the process of movement will leave a substance called Pheromone on its path through the information transferred. Ants can perceive this substance in the course of the campaign as their movement direction.Therefore, a large number of ants showed an information feedback phenomenon: more ants walking on a path, choose the after the greater the probability.

We put on a plane at some point number, and the relationship of the distance between them simplified network chart.If the node can directly reach in a straight line rather than an obstacle, the weights of the edges between them weight of the straight-line distance, otherwise there is no edge between them. As follows:

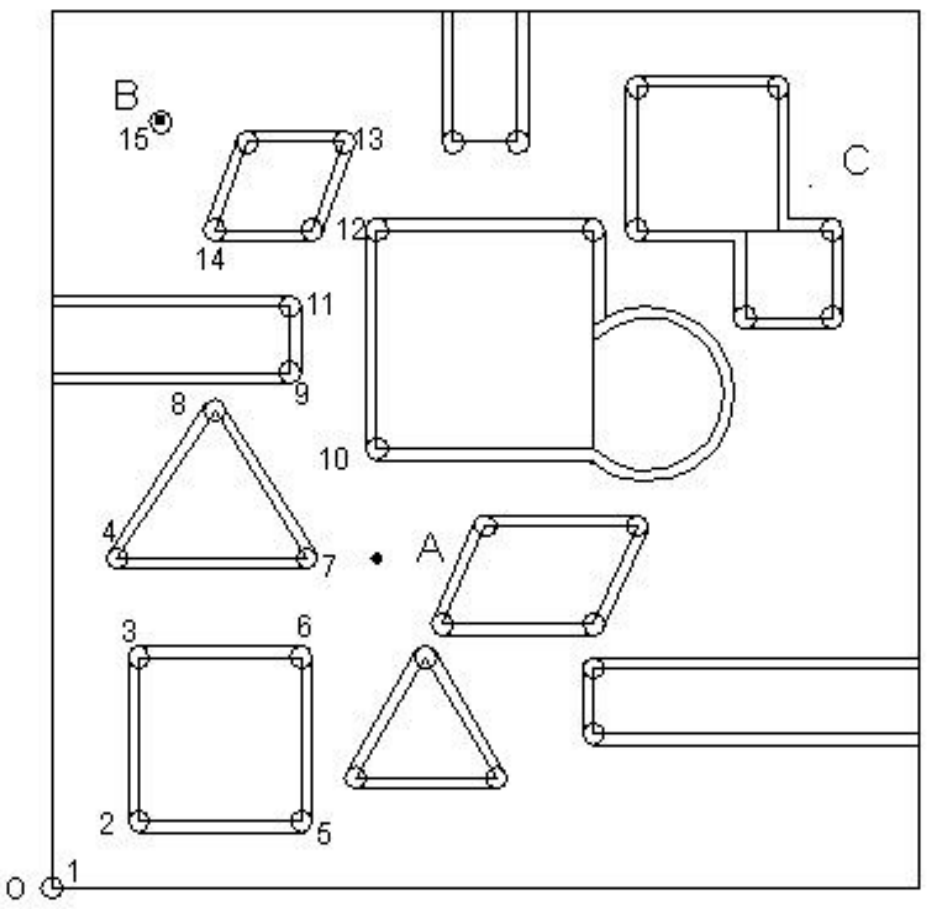

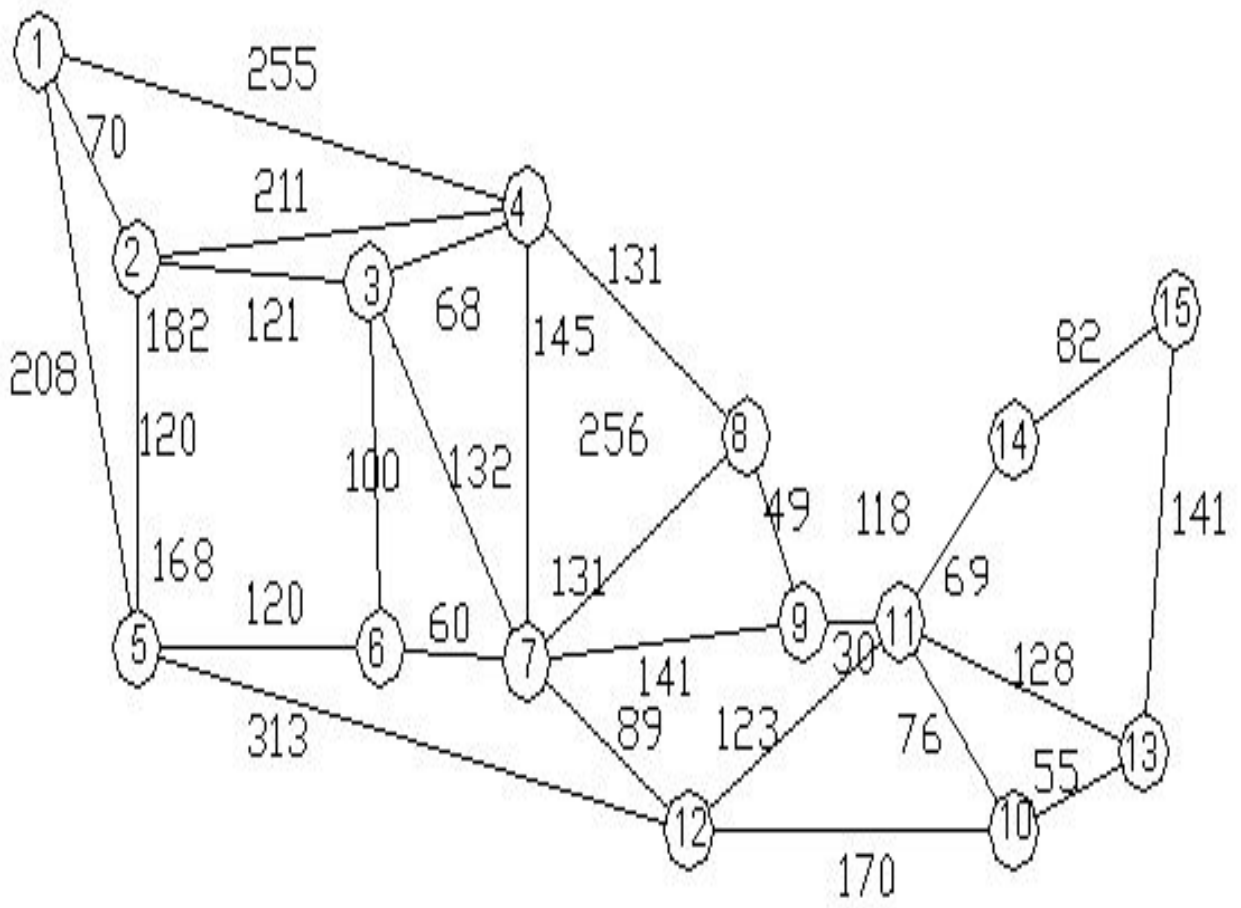

Figure4. Nodes Numbers Figure Figure 5. Empowering network chart

Using ant colony algorithm to select the shortest route from $O \rightarrow B$ from the simplified network chart.

### *1) Ant colony algorithm model*

Value the point $1 \rightarrow 15$ ,0-1wether if it on the path ,form15 bit sequence 0,1, thereby calculating the distance of this path. The distance as a mapping of the pheromone variable, due to the requirements of the most short-circuit, so you can use the countdown or relative distance as the pheromone concentration. Then you get each ant transition probability.If transition probability is greater than the global transfer factor, then the global transfer; otherwise transfer must have step. So that you can step to the global optimal solution close.

### *2) Perform steps*

The first step to initialize N ants. In fact N road, and calculate the current position of ants.

The second step initialization of operating parameters, start the iteration.

The third step in the iterative complement the range of calculated transition probabilities, less than the global transition probability for small-scale search, or a wide range of search.

The fourth step is to update the pheromone, records state, ready for the next iteration.

The fifth step is to enter the third step

The Sixth step output and programming.

### *3) Analysis of results*

The initial state of 50 ants are disorderly distribution, optimized the final position to the polarization, so that we get the optimal solution.

Figure6 are average and optimal curve, from which you can know that the algorithm converges very fast, the effect is better. The shortest path is $1 \rightarrow 4 \rightarrow 8 \rightarrow 9 \rightarrow 11 \rightarrow 14 \rightarrow 15$. Chromosome: 100100011010011, running time is 0.3910.

Therefore, the shortest path from $O \rightarrow B$ as shown below:

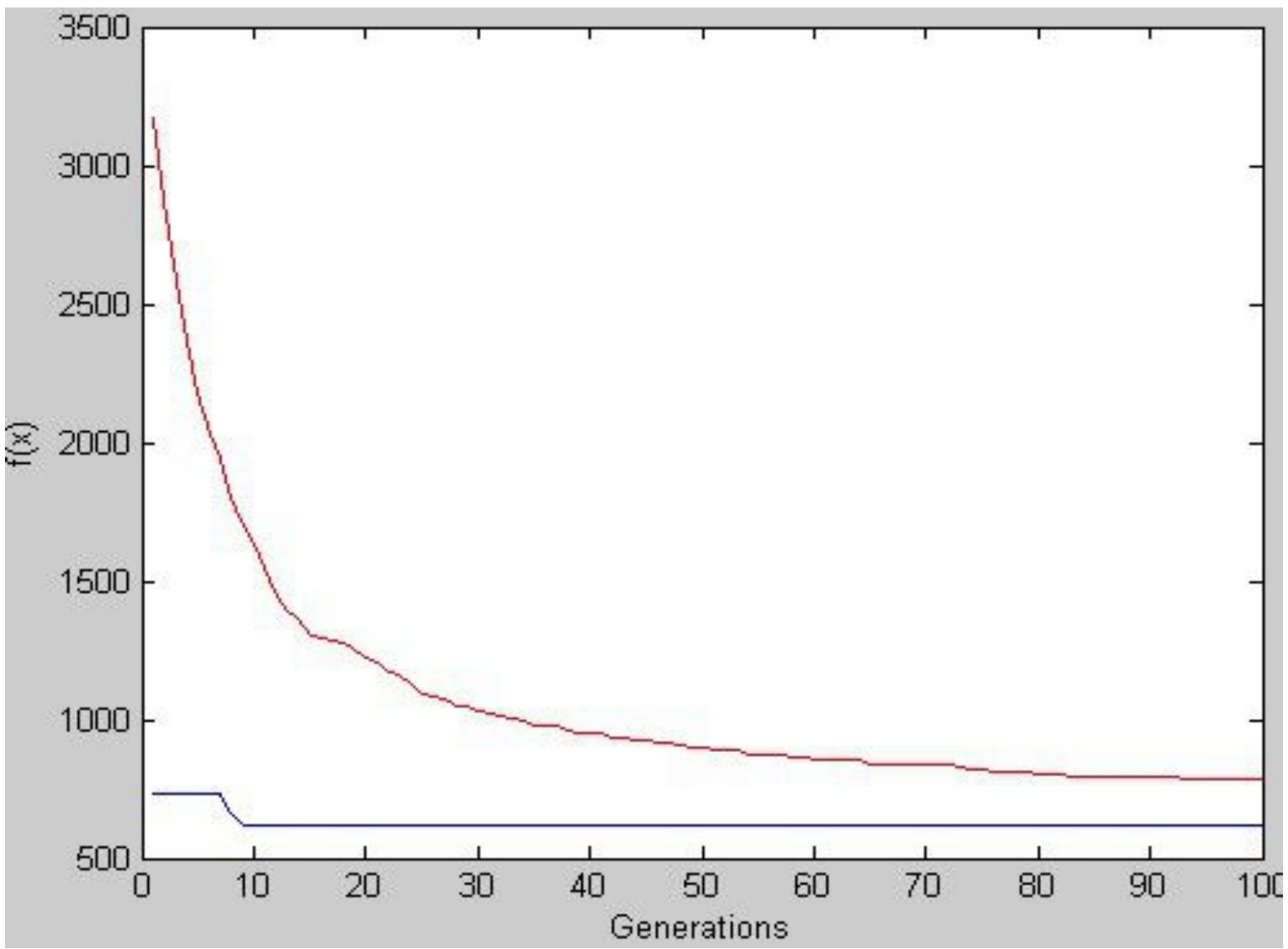

Figure 6. Pheromone concentration average value and the optimum value

## 5.3 The segmented path length from $O \rightarrow B$ optimization model for solving

In the previous section, we have determined to run the route of the shortest node $1 \rightarrow 4 \rightarrow 8 \rightarrow 9 \rightarrow 11 \rightarrow 14 \rightarrow 15$ in Figure 3, but this simplified diagram from only consider a straight line, without regard to the actual deployment of the arc length. Therefore, we put this route segment, making each piece only route to bypass an obstacle.

Based on the above analysis, from this route $O \rightarrow B$ is divided into five sections $L_1(O \rightarrow B_3)$ $L_2(B_3 \rightarrow B_6)$, $L_3(B_6 \rightarrow B_9)$, $L_4(B_9 \rightarrow B_{12})$ $L_5(B_{12} \rightarrow B)$ calculate their length, then the sum thus obtained the shortest path from $O \rightarrow B$.

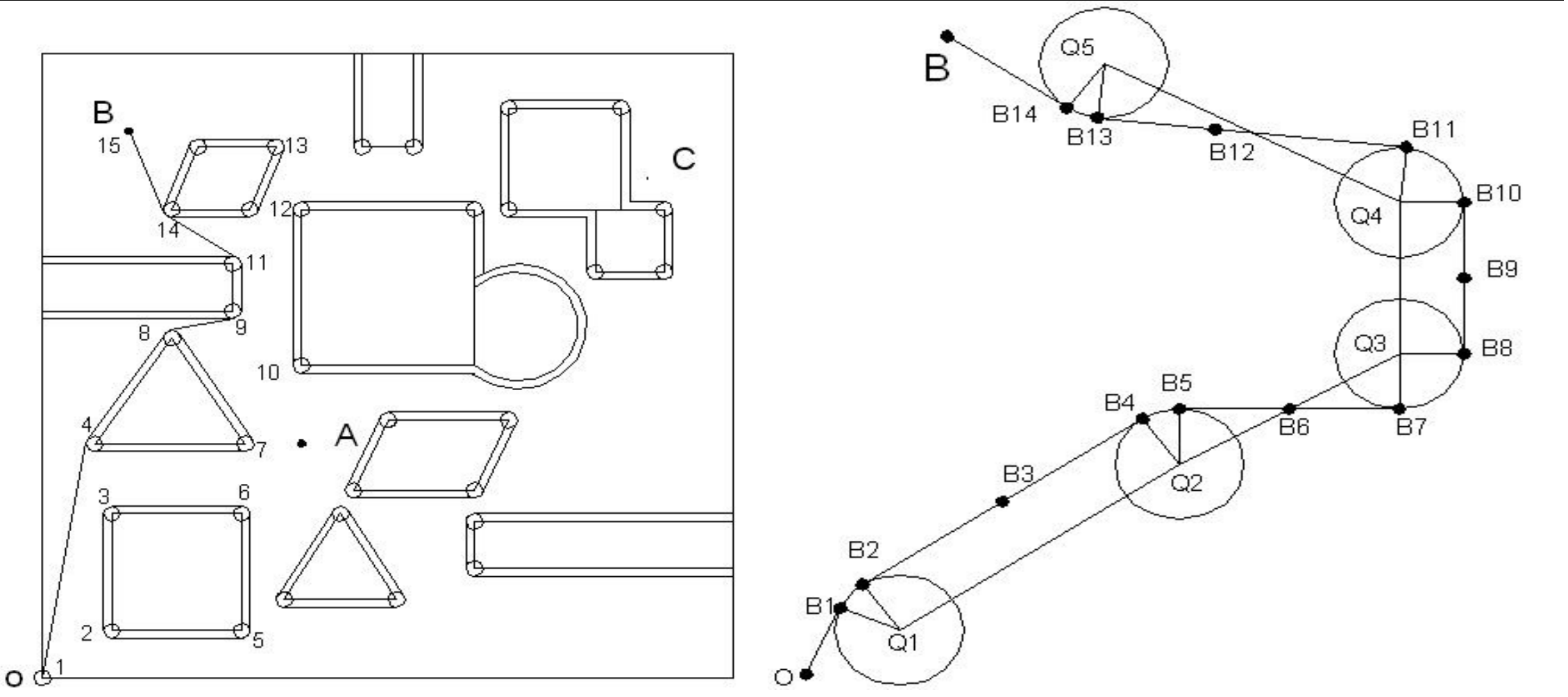

Figure7. Seeking from $O \to B_3$ shortest path

***1) Seeking from $O \to B_3$ shortest path:***

seek the shortest path $O \to A$ structure optimization model is similar to when seeking the shortest path $O \to B_3$, we will coordinate $O$、$B_3$、$Q_1$ as the route start point $(a,b)$, end point $(c,d)$ and the arc center $(m,n)$ coordinates variable values.

Namely: $a=0, b=0; c=100, d=378; m=60, n=300.$

The coordinates of the cut-off point $B_1(x_1,y_1)$, $B_2(x_2,y_2)$ as a decision variable, $O \to B_3$ as the objective function of the length of the shortest structure optimization model as follows:

$$\text{Min } s=\sqrt{(x_1-m)^2+(y_1-n)^2}+\sqrt{(x_1-c)^2+(y_2-d)^2}+20t \tag{6}$$

$$s.t.\begin{cases} x_1 \le m+10 \\ x_1 \ge m \\ y_1 \le n+10 \\ y_1-\sqrt{100-(x_1-m)^2}=n \\ (x_1-a)^2+(y_1-b)^2+100=m^2+n^2 \\ (x_1-c)^2+(y_2-d)^2+100=(m-c)^2+(n-d)^2 \\ \sin t=\dfrac{\sqrt{(x_1-x_2)^2+(y_1-y_2)^2}}{20} \end{cases} \tag{7}$$

Solving the above model, the results are as follows:

The length $OB_3$ of the shortest path: $s_{\min}=397.0986$

Two arc tangent point coordinates:

$$B_1=(50.1353,301.6396)$$
$$B_2=(51.6795,305.547)$$

***2) Seeking $B_3 \to B_6$ shortest path model***

The optimization model with the same route as the previous paragraph, the coordinates of $B_3$、$B_6$、$Q_2$ as the beginning $(a,b)$ of the route, the end $(c,d)$ and the arc center $(m,n)$ coordinates of the variable value. Namely:

$$a = 100, b = 378; c = 185, d = 452.5; m = 150, n = 435.$$

Using the lingo program solving the optimization model, the length of the shortest path. Similarly, we can calculate the shortest path of the other sub-routes. In summary, we have come to the shortest path.

Table4. The result of the length of the shortest path

| **Segm-ented** | **Start** | **End** | **Types of segments** | **Length** |
|---|---|---|---|---|
| 1 | (0,0) | (50.1353,301.6396) | Straight line | 305.7777 |
| 2 | (50.1353,301.6396) | (51.6795,305.547) | （60，300）as the center of the arc | 5.88 |
| 3 | (51.6795,305.547) | (141.6795,440.547) | Straight line | 162.2498 |
| 4 | (141.6795,440.547) | (147.9621,444.79.0) | （150，435）as the center of the arc | 7.7756 |
| 5 | (147.9621,444.79.0) | (222.0379,460.2099) | Straight line | 75.6637 |
| 6 | (222.0379,460.2099) | (230,470) | （220，470）as the center of the arc | 13.6557 |
| 7 | (230,470) | (230,530) | Straight line | 60 |
| 8 | (230,530) | (225.5026,538.3538) | （220，530）as the center of the arc | 9.8883 |
| 9 | (225.5026,538.3538) | (144.5033,591.6462) | Straight line | 96.9536 |
| 10 | (144.5033,591.6462) | (140.6892,596.3523) | （150，600）as the center of the arc | 6.1545 |
| 11 | (140.6892,596.3523) | (100,700) | Straight line | 110.377 |
| Total length | | | | 854.3759 |

## 6. CONCLUSIONS

With the continual development of robotic research in the field of artificial intelligence, the use of ant colony algorithm effectively solves the problem of robot path planning in the practical work of calculation. Our studies show that in a certain range, the optimized model for ant colony algorithm can be used to calculate and design the shortest path when a robot moves from a starting point beyond some obstacles and reaches the specified target points opposite the obstacles without any collision. Nevertheless, further study is necessary in that some limitations still exist in mobile robot path planning via ant colony algorithm, e.g. the model for the shortest path planning remains to be optimized, and whether there are other algorithm solutions to mobile robot path planning etc.

## ACKNOWLEDGMENT

We would like to thank the referees very much for their valuable comments and suggestions, there also got some help from Miss Lu Yan and Ms. Xu Si.

## Appendix：Matlab programming of Ant Colony Algorithm

**function shortroad_ant_main**

```
% Ant main program
clear all;close all;clc;%clear all
tic;%time start
Ant=50;Ger=100;% Running parameter initialization
power=[0 70 1000 276 208 1000 1000 1000 1000 1000 1000 1000 1000 1000 1000
70 0 141 211 120 182 1000 1000 1000 1000 1000 1000 1000 1000 1000
1000 141 0 68 168 100 132 1000 500 1000 1000 1000 1000 1000 1000
276 211 68 0 1000 1000 145 131 1000 1000 1000 1000 1000 1000 1000
208 120 168 1000 0 120 1000 1000 1000 1000 1000 313 1000 1000 1000
1000 182 100 1000 120 0 60 1000 1000 1000 1000 1000 1000 1000 1000
1000 1000 132 145 1000 60 0 131 141 1000 1000 89 1000 1000 1000
1000 1000 500 131 1000 1000 131 0 49 1000 1000 1000 1000 555 1000
1000 1000 1000 1000 1000 1000 141 49 0 1000 30 1000 1000 118 1000
1000 1000 1000 1000 1000 1000 1000 1000 1000 0 76 170 55 1000 1000
1000 1000 1000 1000 1000 1000 1000 1000 30 76 0 123 128 69 1000
1000 1000 1000 1000 313 1000 89 1000 1000 170 123 0 1000 1000 1000
1000 1000 1000 1000 1000 1000 1000 1000 1000 55 128 1000 0 1000 141
1000 1000 1000 1000 1000 1000 1000 555 118 1000 69 1000 1000 0 82
1000 1000 1000 1000 1000 1000 1000 1000 1000 1000 1000 1000 141 82 0];
[PM PN]=size(power);
% Initialization Ant place
v=init_population(Ant,PN);
v(:,1)=1;v(:,PN)=1;% The beginning and end points in the path
% The distance when the information factors concentration
fit=short_road_fun(v,power);
% Distance as small as possible, so and information factors concentration corresponding
T0 = max(fit)-fit;
% Draw the picture
figure(1);grid on;hold on;
plot(fit,'k*');
title('(a) The initial position of ants ');
xlabel('x');ylabel('f(x)');
% Initialization
vmfit=[];vx=[];
P0=0.2;  % P0---- Global transfer factor
P=0.8;  % P ---- Pheromone evaporation coefficient
%C=[];
% Start search the shortest path
for i_ger=1:Ger
 lamda=1/i_ger; % Transfer of step size parameters
        [T_Best(i_ger),BestIndex]=max(T0);% The most information pheromone concentration
 for j_g=1:Ant % To obtain the global transition probability
```

```
 r=T0(BestIndex)-T0(j_g);% The
        best  distance with ant
 Prob(i_ger,j_g)=r/T0(BestIndex);%
        How much speed rate should
        be closer to it
end
for j_g_tr=1:Ant
 if Prob(i_ger,j_g_tr)<P0
        % Local transfer ---- Small step
        transfer
  M=rand(1,PN)<lamda;
  temp=v(j_g_tr,:)-
        2.*(v(j_g_tr,:).*M)+M;
 else
         %Globle transfer----
        Big step transfer
  M=rand(1,PN)<P0;
  temp=v(j_g_tr,:)-
        2.*(v(j_g_tr,:).*M)+M;
 end
        % Reput the beginning and end
        points,which is not in
        the moving process of change
 temp(:,1)=1;temp(:,end)=1;
 if
        short_road_fun(temp,power)<s
        hort_road_fun(v(j_g_tr,:),powe
        r)
         %Recorder
  v(j_g_tr,:)=temp;
 end
end
% Update the information factors，
        To prepare for the
        next iteration
fit=short_road_fun(v,power);
T0 = (1-P)*T0+(max(fit)-fit);%
        information factors evaporation
[sol,indb]=min(fit);
v(1,:)=v(indb,:);%
        Record the iteration of the state
media=mean(fit);
vx=[vx sol];
 vmfit=[vmfit media];
end
% % % % % % % % % % % % % %
        % % % % % % % % % % %
        % % % % % % % %
%%%% The last result
disp(sprintf('\n')); % Blank one line
% Display the optimal solution and
        the optimal value
disp(sprintf('Shortroad is
        %s',num2str(find(v(indb,:)))));
        %num2strData into character.
disp(sprintf('Mininum is %d',sol));
v(indb,:)
% Graphic display optimal results
figure(2);grid on;hold on;
plot(fit,'r*');
title('The last place of Ant');
xlabel('x');
ylabel('f(x)');
% Graphic display optimal and
        average function value trend
figure(3);
plot(vx);
title(' optimal, The average function
        value trend ');
xlabel('Generations');ylabel('f(x)');
hold on;plot(vmfit,'r');hold off;
runtime=toc%Time end
end
%%
function fit=short_road_fun(v,power)
[vm vn]=size(v);
fit=zeros(vm,1);% Record  distance
        of each path
for i=1:vm
 I=find(v(i,:)==1);% Looking for
        the point on the path
 [Im,In]=size(I);
 for j=1:In-1
  fit(i)=fit(i)+power(I(j),I(j+1));%
        Find distance of the path
 end
```

```
end
end
%%
%Function init_population
function v=init_population(n1,s1)
v=round(rand(n1,s1));% Initializes all
    the Ants
END
```

**Authors**

**(1) GUO Yue**

PhD in Management Science and Engineering, MAIB, BBA

Profersor, Dean of Department of Enterprising Management

Ningbo University of Technology

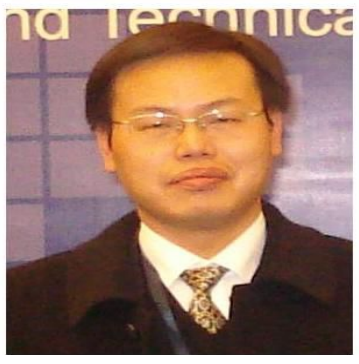

In 2010 Dr. Guo award Chinese Provincial New Century Hundred-Thousand-Ten thousand Engineering Talent, in 2011 he also got Chinese Provincial Backbone Middle-Young Aged Teachers. Dr. Guo has been published over 50 papers and 8 books in his academic field. In management science and engineering field he also do some consulting services to solve the actual problems for local companies and enterprises.

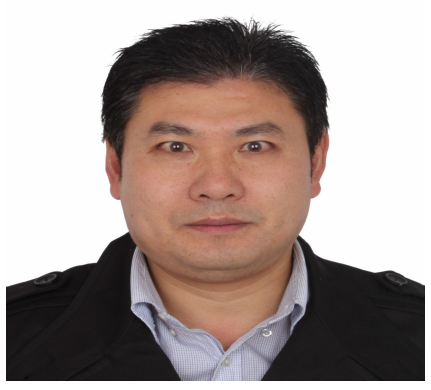

**(2) SHEN Xuelian**

PhD graduated from Southwest University of Finance and Economics, as a lecturer teaching in economics field courses at Ningbo University of Technology. He got his master degree from Swiss University.

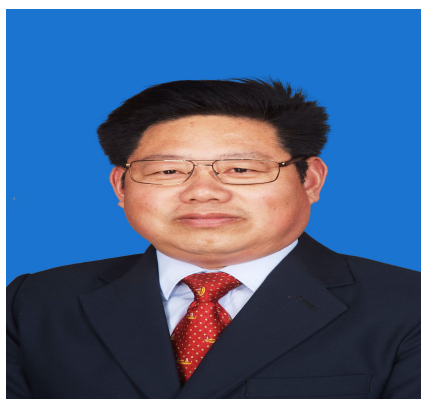

**(3) ZHU Zhanfeng**

Postdoctoral fellow, PhD in Management Science and Engineering, MBA, BA

Profersor, Headof Department of Economics & Management

Ningbo University of Technology

He was visiting scholar of University of magdeburg in Germany; and got advanced logistics management division, registered senior consultant and Doctoral tutor. Also presently for ningbo college of engineering management, and director of the institute of the ningbo college of engineering development of small and medium-sized enterprises. At the same time, he was also the ministry of education of institutions of higher learning MBA class teaching steering committee of logistics and electronic commerce, deputy director of the national demonstration on repository construction project logistics management specialty teaching work committee.